\def\year{2021}\relax
\documentclass[letterpaper]{article} 
\usepackage{aaai21}  
\usepackage{times}  
\usepackage{helvet} 
\usepackage{courier}  
\usepackage[hyphens]{url}  
\usepackage{graphicx} 
\usepackage{natbib}  
\usepackage{caption} 
\usepackage{csquotes}
\usepackage{amsmath}
\usepackage[ruled, linesnumbered]{algorithm2e}
\usepackage{soul}
\newcommand{\xAI}{xAI-GAN}

\title{\xAI{}: Enhancing Generative Adversarial \\Networks via Explainable AI Systems}
\author {
        Vineel Nagisetty\textsuperscript{\rm 1},
        Laura Graves\textsuperscript{\rm 1},
        Joseph Scott\textsuperscript{\rm 1} and
        Vijay Ganesh\textsuperscript{\rm 1} \\
}
\affiliations {
    \textsuperscript{\rm 1} University of Waterloo\\
    \{vineel.nagisetty, laura.graves, joseph.scott and vijay.ganesh\}@uwaterloo.ca
}
\begin{document}

\maketitle

\begin{abstract}
Generative Adversarial Networks (GANs) are a revolutionary class of Deep Neural Networks (DNNs) that have been successfully used to generate realistic images, music, text, and other data. However, GAN training presents many challenges, notably it can be very resource-intensive. A potential weakness in GANs is that it requires a lot of data for successful training and data collection can be an expensive process. Typically, the corrective feedback from discriminator DNNs to generator DNNs (namely, the discriminator's assessment of the generated example) is calculated using only one real-numbered value (loss). By contrast, we propose a new class of GAN we refer to as \xAI{} that leverages recent advances in explainable AI (xAI) systems to provide a \enquote{richer} form of corrective feedback from discriminators to generators. Specifically, we modify the gradient descent process using xAI systems that specify the reason as to why the discriminator made the classification it did, thus providing the \enquote{richer} corrective feedback that helps the generator to better fool the discriminator. Using our approach, we observe \xAI{}s provide an improvement of up to 23.18\% in the quality of generated images on both MNIST and FMNIST datasets over standard GANs as measured by Fréchet Inception Distance (FID). We further compare \xAI{} trained on 20\% of the data with standard GAN trained on 100\% of data on the CIFAR10 dataset and find that \xAI{} still shows an improvement in FID score. Further, we compare our work with Differentiable Augmentation - which has been shown to make GANs data-efficient - and show that \xAI{}s outperform GANs trained on Differentiable Augmentation. Moreover, both techniques can be combined to produce even better results. Finally, we argue that \xAI{} enables users greater control over how models learn than standard GANs.
\end{abstract}

\section{Introduction}
Generative Adversarial Networks (GANs), introduced only a few short years ago, already have had a revolutionary impact on generating data of varied kinds such as images, text, music, and videos~\cite{original_gan}. The critical insight behind a GAN is the idea of corrective feedback loop from a deep neural network (DNN) called the \textit{discriminator} back to a \textit{generator}. However, a notable weakness of GANs is that they require a lot of data for successful training. For example, in order to learn to write digits, a human may only need a few ($\le$10) examples before she or he learns to replicate them whereas a GAN would often need several orders of magnitude more data ($\ge$1000s) to learn the same task.

The collection of high-quality labelled data for GANs and other machine learning algorithms often requires a lot of resources such as time and money and is considered a `major bottleneck' in machine learning research~\cite{data}. As a consequence, there is a need for finding other ways of making GANs more data-efficient. Observe that, in the standard GAN architecture, the feedback provided by the discriminator to the generator is calculated using only one value (loss). This feedback is calculated as follows. The discriminator takes as input data from the generator to make a prediction. Next, loss is calculated based on this prediction and is used by the discriminator to provide feedback to the generator. This feedback is then used by the generator to perform gradient descent and update it's parameters so as to better fool the discriminator. The feedback provided thus originates from this single real-numbered value (loss). Hence, the research questions we address in this paper are the following: is it possible to provide \enquote{richer} corrective feedback from the discriminator back to the generator? If so, does it enable GANs to be more data-efficient?

\subsection{\bf An Overview of \xAI{}} 
\label{overview}
\begin{figure}[t]
\centering
\includegraphics[width=0.9\columnwidth]{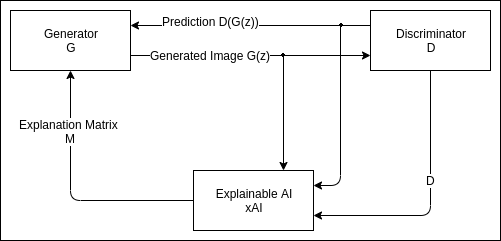}  
\caption{System Architecture of \xAI{}.} 
\label{fig:architecture}
\end{figure}

To answer the above-mentioned research questions, we propose a new class of GANs we refer to as \xAI{} wherein it is possible to provide \enquote{richer} corrective feedback (more than a single value) during training from discriminator to generators via Explainable AI (xAI) systems. A high-level system architectural overview of our \xAI{} system is given in Figure~\ref{fig:architecture}. Consider the problem of training a GAN with the aim of producing images of digits. Initially, the untrained generator $G$ is given a noise sample $z$ from a prior noise distribution $p_z$ and produces an example $G(z)$ that is then given to discriminator $D$. The loss is calculated, and then the generated image $G(z)$, the discriminator $D$, and the output of the discriminator $D(G(z))$ are fed into an xAI system that produces an explanation as to why the image resulted in that loss. This explanation is then used to guide the training of the generator (refer Section~\ref{sec:overview_xAI} for details).

A common analogy for GAN training is that of a counterfeiter (the generator) and a detective (the discriminator) playing an adversarial game where the counterfeiter makes a fake and the detective tries to tell if it's real or not. Over the training process, the detective and counterfeiter both get better at their jobs, with the end goal being that the counterfeiter is so proficient that their fakes can pass for the real thing. To extend this analogy to the \xAI{} setting, our method works by using an expert in the field (the xAI system) to help improve the counterfeiter. When the detective recognizes a fake, the expert tells the counterfeiter what parts of the fake tipped off the detective. The counterfeiter is thus able to learn better why the detective detected a fake, and make better decisions to avoid pitfalls in future training.

\noindent{\bf Explanation Matrix:} The explanation produced by xAI systems are converted to the form of an \enquote{explanation matrix} $M$, wherein, for every feature in an example (e.g., an input image), a value in the range [0,1] is assigned to the corresponding cell of the matrix $M$. If a feature (more precisely, the corresponding cell in $M$) is assigned value 0 (or close to 0), then it means that pixel had no impact on the classification decision made by the discriminator $D$. If a feature is assigned a value of 1 (or close to 1), then that means that feature is very important. This could be due to the feature being \enquote{determinative} in the classification made by $D$ or when it \enquote{hurts} the classification made by $D$ (more precisely, if the feature were to be changed, then the confidence of $D$ in its classification would improve). Creating the explanation matrix in this manner helps us focus the learning process on the most influential features, regardless of whether those features were beneficial or harmful to the classification.

\noindent{\bf xAI-guided Gradient Descent:} The matrix $M$ generated by the xAI system is then used in a modified gradient descent algorithm (see Algorithm~\ref{alg:GENTRAINING}) to update the weights of the generator as follows: traditionally, in a GAN the weights of the generator are modified by first computing the gradient of generator's output with respect to the loss and then applying the chain rule. We modify this algorithm by first computing the explanation matrix $M$ (via the xAI system) and then calculating the product (specifically, an element wise or a Hadamard product~\cite{hadamard}) between $M$ and the gradient of the generator output with respect to the loss $\Delta_{G(z)}$. More precisely, the explanation matrix $M$ is used to mask the gradient function, and consequently the \enquote{importance of the pixels} that went into the discriminator's classification are taken into account in the modification of the generator's weights during the application of the gradient descent. 

Using our approach, one can foresee that users might be able to augment such explanation matrices with specifications that spell out relationships (using logical formulas) between the update methods for the various weights of a generator. We would like to emphasize that the standard gradient descent method simply moves toward the greatest decrease in loss over an $n$-dimensional space, while by contrast, xAI-guided gradient descent algorithms can give users greater control over the learning process.

\subsection{Contributions}
\begin{enumerate}
    \item {\bf xAI-guided Gradient Descent Algorithm and \xAI{}:} Our key contribution is an xAI-guided gradient descent method (and a resultant GAN we refer to as \xAI{}) that utilizes xAI systems to focus the gradient descent algorithm on weights that are determined to be most influential by the xAI system (refer Section~\ref{sec:xAI}). We implement several different versions of \xAI{} using 3 different state-of-the-art xAI systems~(refer Section~\ref{sec:xAI}), namely saliency map~\cite{saliency}, shap~\cite{shap_paper}, and lime~\cite{lime}. In Section~\ref{sec:control}, we discuss how xAI-guided gradient descent methods can give those training models greater control over the learning process.
    
    \item {\bf Experimental Evaluation of Quality of Images produced by \xAI{} vs. Standard GAN:} We performed experiments to evaluate the quality (as measured by Fréchet Inception Distance, abbreviated as FID~\cite{fid}) of \xAI{}s relative to standard GANs. We show that on MNIST and Fashion MNIST datasets, \xAI{}s achieve an improvement of up to 23.18\% in FID score compared to standard GANs (refer Section~\ref{sec:mnist}).

    \item {\bf Experimental Evaluation of Data Efficiency of \xAI{}s vs. Standard GANs:} We extend our experiment to the CIFAR 10 dataset, using only 20\% of the data for \xAI{} while letting standard GAN use 100\% of the data. We show that \xAI{} outperforms standard GAN in FID score even in this setting. We further compare our work with Differentiable Augmentation~\cite{data_efficient_gan} technique which has been shown to improve data-efficiency of GANs. We show that \xAI{} outperforms Differentiable Augmentation, resulting in a better FID score. Finally, we modify our \xAI{} to incorporate Differentiable Augmentation and show that the resulting model has better performance than either version. 
\end{enumerate}

\section{Related Work}
Goodfellow et al. were the first to introduce GANs in~\shortcite{original_gan}. Since then, GANs have continued to be a popular research topic with many versions of GANs developed~\cite{gan_survey}. GANs can be broadly classified based on their architecture~\cite{dc_gans,cgans,infogan,adv_gan} and the type of objective function used~\cite{unrolled_gan,wasserstein,lsgan,mmdgan,energyGAN,wasserstein_improved}. To the best of our knowledge, there is no GAN that uses xAI feedback for training, thus making \xAI{} the first of its kind. We note that the xAI-guided gradient descent algorithm is independent of architecture or type of objective function used, and therefore can be applied to make any type of GAN an \xAI{}.

Differentiable Augmentation~\cite{data_efficient_gan} is a recent technique that aims to make GANs more data-efficient by augmenting the training data. The main idea behind this technique is to increase the data via various types of augmentations on both real and fake images during GAN training. These augmentations are differentiable and so the feedback from the discriminator can be propagated back to the generator through the augmentation. On the other hand, while \xAI{} also aims to make GANs more data-efficient, this is done by passing \enquote{richer} information from the discriminator to the generator through an xAI system. We compare \xAI{}s with Differentiable Augmentation in section~\ref{sec:cifar} and show that they can be combined to provide further improvements in data-efficiency.

ADAGRAD~\cite{adagrad} is an optimization algorithm that maintains separate learning-rates for each parameter of a DNN based on how frequently the parameter is updated. On the other hand, \xAI{} uses xAI feedback to determine how generator parameters are updated (refer Section~\ref{sec:xAI}). 

\noindent{\bf Explainable AI (xAI) Systems:} As AI models become more complex, there is an increasing demand for {\it interpretability or explainability} of these models from decision makers, stakeholders, and lay users. In addition, one can make a strong case for a scientific need for explainable AI. Consequently, there has been considerable interest in xAI systems aimed at creating interpretable AI models that enable human understanding of AI systems~\cite{explainable_survey}.

One way to define xAI systems is as follows: they are algorithms that, given a model and a prediction, assigns values to each feature of an input that measures how important that feature is to the prediction. There have been several different xAI systems applied to DNNs with the goal of improving our understanding of how these systems learn. These systems can do so in a variety of ways, and approaches have been developed such as ones using formal logic~\cite{ignatiev2019abduction}, game-theoretic approaches~\cite{lundberg2017unified}, or gradient descent measures~\cite{shrikumar2017learning}. These systems output explanations in a variety of forms such as ranked lists of features, select subsets of the feature sets, and values weighting the importance of features of input data used to train machine learning models.

\section{The xAI Systems}
We implement and compare several xAI systems. Note that our \xAI{} is xAI system agnostic and can be used with any xAI system. However, the efficacy of the training of \xAI{} depends on the efficacy of the xAI system and hence selecting the appropriate xAI system is crucial. 

\subsection{Saliency Map}
Inspired by the processes by which animals focus attention, saliency maps~\cite{saliency} compute the importance of each feature in a given input to the resulting classification by a DNN model. In order to compute a saliency map, a DNN model $M$, image $x$ and target label $y$ are required. The loss of the prediction $M(x)$ is computed with respect to $y$ and used to perform backpropagation to the calculate the gradient $\nabla x$. This is then normalized to produce the saliency map. While there are similarities between saliency maps and the process by which the gradients passed by the discriminator to the generator is calculated, there are some differences. Notably, in the case of color images (such as CIFAR10 dataset), saliency map computes the maximum magnitude of $\nabla x$ for each pixel across all color channels, while the gradients are computed for each color channel separately.

\subsection{Lime}
Lime~\cite{lime}, short for Local Interpretable Model-Agnostic, is used to explain the predictions of a ML classifier by learning an interpretable local model. Given a DNN model $M$ and input $x$, Lime creates a set of new $N$ inputs $x_1,...,x_N$ by slightly perturbing $x$. It then queries $M$ on these new inputs to generate labels $y_1 ,...,y_N$. The new inputs and labels are used to train a simple regression model which is expected to approximate $M$ well in the local vicinity of $x$. The weights of this local model are used to determine the feature importance of $x$.


\subsection{DeepSHAP} DeepSHAP~\cite{lundberg2017unified} is a combination of the DeepLIFT platform and {\it Shapley value explanations}. Introduced in  2017, the platform is well-suited for neural network applications and is freely available. DeepSHAP is an efficient Shapley value estimation algorithm. It uses linear composition rules and backpropagation to calculate a compositional approximation of feature importance values.

\noindent{\bf Shapley Value Estimation:} Classic Shapley regression values are intended for linear models, where the values represent feature importance. Values are calculated by retraining models on every subset of features $S \subseteq F$ and valuing each feature based on the prediction values on models with that feature and without. Unfortunately, this method not only requires significant retraining but also requires at least $2^{|F|}$ separate models to cover all combinations of included features. Methods to approximate the Shapley values by iterating only over local feature regions, approximating importance using samples from the training dataset, and other approaches have been proposed to reduce computational effort. 

\noindent{\bf The DeepLIFT xAI system:} DeepLIFT uses a set of reference inputs and the consequent model outputs to identify the importance of features~\cite{shrikumar2017learning}. The difference between an output and a reference output, denoted by $\Delta y$, is explained in terms of the differences between the corresponding input and a reference input, given by $\Delta x_i$. The reference input is chosen by the user based on domain knowledge to represent a typical uninformed state. It is often a set of images from the original dataset that the model is trained on. Each feature $x_i$ is given a value $C_{\Delta x_i \Delta y}$ which measures the effect of the model output on that feature being the reference value instead of its original value. The system uses a summation property where the sum of each feature's changes sum up to the change in the model output $\Delta _o$ of the original in comparison to the reference model: $\sum_{i=1}^{n} C_{\Delta x_i \Delta y} = \Delta _o$.

\section{Detailed Overview of \xAI{} Systems}
\label{sec:overview_xAI}

\begin{algorithm}[t!]
    \SetAlgoLined
    \SetKwInOut{Input}{input} \SetKwInOut{Output}{output}
    \Input{generator G}
    \Input{discriminator D}
    \Input{boolean Flag use\_xAI}
    \Output{trained generator G}
    \ForEach{noise sample $z$}{
    Loss L = $Loss(1 - D(G(z))$\\
    compute Discriminator Gradient $\Delta_{D}$ from L\\
    compute Generated Example Gradient $\Delta_{G(z)}$ from $\Delta_{D}$\\
    \uIf{use\_xAI is True}{
    compute Explanation Matrix $M$ using xAI \\
    compute Modified Gradient $\Delta_{G(z)}'$ = $\Delta_{G(z)} + \alpha * \Delta_{G(z)} * M$\\
   compute Generator Gradient $\Delta_{G}$ from $\Delta_{G(z)}'$\\
   }
    \uElse{ compute Generator Gradient $\Delta_{G}$ from $\Delta_{G(z)}$\\
    }
    }
    update Generator parameters $\theta_{G}$ using $\Delta_{G}$
    \caption{Generator Training Algorithm. Note that the code block under the if \textit{use\_xAI} block only applies to the xAI-guided generator training.}
    \label{alg:GENTRAINING}
\end{algorithm}

In this section, we provide a detailed overview of our \xAI{} system (please refer to Figure~\ref{fig:architecture} for the system architecture of \xAI{} and Algorithm~\ref{alg:GENTRAINING} for the gradient descent algorithm for generator training) and contrast it with standard GAN architectures as well as the way they are trained. The intuition behind the xAI-guided generator training process is that the xAI system acts as a guide, shaping the gradient descent in a way that focuses generator training on input features the discriminator recognizes as important.

\subsection{Generator Training in Standard GANs}

Briefly, standard GAN architectures consist of a system of paired DNNs, namely, a discriminator $D$ and a generator $G$. The standard training method involves alternate cycles of discriminator and generator training. Initially, the discriminator is trained on a mini batch of examples drawn from both training data from the target distribution, as well as data generated by the untrained generator (which initially is expected to be just noise). These examples are correctly labeled as \enquote{real} (if they were from the training set) or \enquote{generated} (if they are from the generator). 

Subsequently, the generator is trained as follows (please refer to Algorithm~\ref{alg:GENTRAINING}): a selection of noise samples are drawn from the noise prior and passed through the generator to get a batch of generated examples (line 1). This batch is labeled as \enquote{real} and given to the discriminator, where the loss is found (line 2), and then used to update the generator parameters (the corrective feedback step). More precisely, the discriminator's gradient $\Delta_{D}$ is computed using the parameters of the discriminator and its loss (line 3), which is used to find the gradient of the generated example $\Delta_{G(z)}$ (line 4). Further, the gradients of all layers in the generator $\Delta_{G}$ are then computed using $\Delta_{G(z)}$ (line 10). Finally, the parameters $\Theta_{G}$ of the generator are updated using $\Delta_{G}$ (line 12) - completing one training iteration.

In subsequent iterations, the discriminator receives mini batches of real and generated examples from the generator trained in the previous iterations. The ideal termination condition for this process is when both the generated examples are high-quality and the discriminator is unable to distinguish between \enquote{real} and \enquote{generated} examples. 


\begin{table}[tb]
    \centering
    \begin{tabular}{|c|c|}
    \hline
    \textbf{Generator} & \textbf{Discriminator} \\ \hline
    Dense (100, 256) & Dense (1024, 1296) \\ \hline
    Dense (256, 512) & Dense (1296, 512) \\ \hline
    Dense (512, 1296) & Dense (512, 256) \\ \hline
    Dense (1296, (32, 32)) & Dense (256, 1) \\ \hline
\end{tabular}
    \caption{Model Architecture of Fully Connected GAN used in the context of MNIST and Fashion MNIST experiments.}
    \label{tab:arch_fcgan}
\end{table}

\begin{table}[tb]
    \centering
    \begin{tabular}{|c|c|}
    \hline
    \textbf{Generator} & \textbf{Discriminator} \\ \hline
    Conv (4 x 4) & Conv (16 x 16) \\ \hline
    Conv (8 x 8) & Conv (8 x 8) \\ \hline
    Conv (16 x 16) & Conv (4 x 4) \\ \hline
    Conv (32 x 32) & Conv (1 x 1) \\ \hline
\end{tabular}
    \caption{Model Architecture of DC-GAN which we used in the context of the CIFAR10 experiments.}
    \label{tab:arch_dcgan}
\end{table}

\subsection{xAI-guided Generator Training in \xAI{}s}
\label{sec:xAI}

We start our description of xAI-guided training by first observing that in the standard GAN setting the discriminator calculates the corrective feedback to the generator using only a single value (loss) per generated image. The entire point of xAI-guided training is to augment this feedback with the \enquote{reason} for the discriminator's decision, as determined by the xAI system. 

During our \textit{xAI-guided gradient descent} generator training process, the backpropagation algorithm is modified to focus generator training on the most meaningful features for the discriminator's prediction (please refer to lines 5-8 of Algorithm~\ref{alg:GENTRAINING}). Following with propagating the loss through the discriminator to find $\Delta_G(z)$, we use an xAI system $E$ to find $M = E(G(z))$ (line 6). $M$ is a set of real values $\in [0,1]$, where greater values represent features that are more important to the discriminator's prediction. The Hadamard (element wise) product of $\Delta_{G(z)}$ and $M$ is calculated to get the modified gradient $\Delta_{G(z)}'$ (line 7). In an intuitive sense, the explanation $M$ acts as a mask for $\Delta_{G(z)}$, focusing the gradient on the most important features and limiting the gradient on the less important ones. From there, the gradients of the generator $\Delta_G$ are calculated from $\Delta_{G(z)}'$ using a small value for $\alpha$ (line 8) and the parameters are then updated (line 12).

\subsection{\xAI{} Implementation Details}

We implemented\footnote{The code of our implementation can be found at: https://github.com/explainable-gan/XAIGAN} \xAI{} using Pytorch 1.6~\cite{pytorch}, an open source machine learning framework popular in deep learning research. For saliency and shap xAI systems we used Captum 0.2.0~\cite{captum}, an open source interpretability framework developed by the team at Pytorch. For the lime-based \xAI{} system we use the implementation from Lime 0.2.0.1~\cite{lime}. We process the explanation matrix M generated by each of the xAI systems by taking the absolute value and normalizing the matrix to create a mask vector with values in range $[0,1]$. 

Pytorch notably has the \texttt{autograd}~\cite{pytorch_autograd} package which handles automatic differentiation of all tensors. In order to provide xAI-guided feedback to the generator, we overrode the \texttt{register\_backward\_hook} function normally used to inspect gradients. We modified the gradients of the output layer of the generator using the resultant vector computed by the Hadamard (element wise) product with the computed mask. This modified gradient is back-propagated through the generator via the \texttt{autograd}. After extensive testing, we found that switching on the xAI-guided gradient descent after half the number of training epochs gives the best results. This is because the discriminator would have learnt the distribution of the task at hand to a certain extent and consequently the xAI system is likely to produce better explanations.

\section{Experimental Results}

\label{sec:experiments}
\begin{figure}[tb]
\centering
\includegraphics[width=0.9\columnwidth]{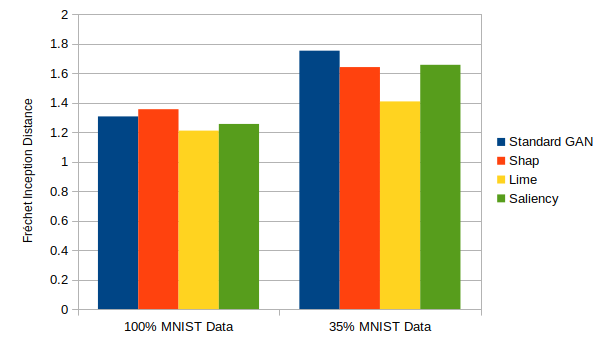}
\caption{FID scores on MNIST Dataset}\label{fig:results_mnist}
\vskip 4mm
\includegraphics[width=0.9\columnwidth]{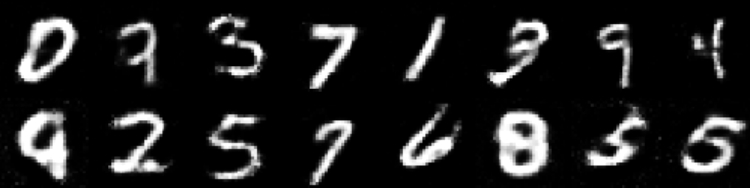}
\caption{Sample of Images Generated by the \xAI{}$_{lime}$ System using 35\% Data on the MNIST Dataset}
\label{fig:viewMNIST}
\end{figure}

We performed extensive experimental evaluation of our \xAI{} implementation that used 3 different xAI systems, comparing against standard GAN. We performed these experiments on three different datasets:
\begin{enumerate}
    \item MNIST~\cite{mnist} a collection of 70,000 28x28 grayscale images of handwritten digits,
    \item Fashion MNIST~\cite{fmnist} a collection of 70,000 28x28 grayscale images of clothing, and
    \item CIFAR10~\cite{cifar10} a collection of 60,000 3x32x32 color images of objects.
\end{enumerate}

For the MNIST and Fashion MNIST datasets, we resized the images to 32x32 and used fully connected GANs for both standard and \xAI{}c the architecture for which is shown in Table~\ref{tab:arch_fcgan}. Leaky relu was the activation used for all but the last layers in the generator and discriminator. In the last layer, we used tanh for the generator, and sigmoid for the discriminator. A dropout rate of 0.3 was used during training in discriminator. For CIFAR10 dataset, we use a DC-GAN architecture for both standard and \xAI{} as described in Table~\ref{tab:arch_dcgan}. The generator and discriminator use four convolutional layers, each with a stride of 2 and padding of 1. Each of them also use a batchnorm layer after every convolutional layer, except for the last one. The activation functions are identical to the fully-connected GAN architecture. 

\subsection{Experimental Setup} 

\begin{figure}[tb]
\centering
\includegraphics[width=0.9\columnwidth]{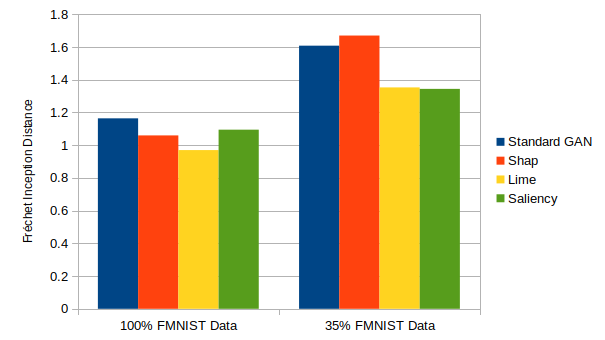}
\caption{FID scores on Fashion MNIST Dataset }\label{fig:results_fmnist}
\vskip 4mm
\includegraphics[width=0.9\columnwidth]{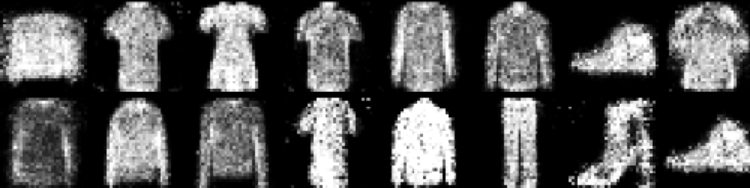}
\caption{Sample of Images Generated by the \xAI{}$_{shap}$ System using 35\% Data on the Fashion MNIST Dataset}
\label{fig:viewFMNIST}
\end{figure}

The batch size was selected to be 128 in our experiments. The Adam optimizer~\cite{adam} was used for both generator and discriminator training. We used a learning rate of 0.0002. We ran experiments using Amazon's EC2 on a p2.xlarge instance which uses 1 Nvidia’s K80 GPU with 64GiB RAM.

\subsection{Evaluation Criteria}
Based on a thorough literature survey of metrics~\cite{gan_survey} for the image domain, we developed the following criteria in order to perform a fair comparison of \xAI{}s vs. standard GANs:

\begin{enumerate}
    \item \textbf{Fréchet Inception Distance (FID):}
    We opted to use Fréchet Inception Distance (FID) to measure quality since it has been shown to be consistent with human evaluation of quality~\cite{fid}. FID was introduced by Heusel et al., to address the shortcomings of Inception Score (IS) such as the latter's inability to detect intra-class mode dropping and vulnerability to noise~\shortcite{fid}. 
    
    At a high level, FID converts a set of images to the feature space provided by a specific layer in the Inception model. Various statistics, such as the mean and covariance are computed on the activation values of that layer to generate a multi-dimension Gaussian distribution. Finally, the Fréchet distance of the two distributions created using the generated and the training images is computed and provided as the output. In order to apply FID to MNIST and Fashion MNIST, we use the LeNet classifier, consistent with~\cite{kid}.
    
    \item \textbf{Training Time:} We also measure the time required for training to identify the overhead added by xAI systems.
\end{enumerate}  

\subsection{Results on MNIST and Fashion MNIST}
\label{sec:mnist}

\begin{table}[bt]
    \centering
    \begin{tabular}{|c|c|c|c|c|}
    \hline
      Dataset & Standard & Shap & Lime & Saliency\\ \hline
      MNIST & 1221.19 & 9925.1 & 38865.93 & 2215.73 \\ \hline
      FMNIST & 991.11 & 9694.81 & 39237.33 & 2162.24\\ \hline
    \end{tabular}
    \caption{Average experiment time taken (in seconds) on MNIST and Fashion MNIST Datasets}
    \label{tab:times_mnist}
\end{table}

\begin{table}[tb]
    \centering
    \begin{tabular}{|c|c|c|c|c|}
    \hline
      Dataset & Standard & Shap & Lime & Saliency\\ \hline
      CIFAR10 & 536.76 & 13290.34 & 12988.32 & 1214.44 \\ \hline
      +Diff & 674.42 & 14319.87 & 13352.37 & 1175.99 \\ \hline
    \end{tabular}
    \caption{Average experiment time taken (in seconds) by each GAN on MNIST and Fashion MNIST (FMNIST) datasets.}
    \label{tab:times_cifar}
\end{table}

We ran the experiments on both MNIST and Fashion MNIST datasets using two settings: 100\% data and 35\% data (to see the performance of \xAI{} when data is scarce). The results of the experiments on MNIST dataset can be found in Figure~\ref{fig:results_mnist}. For 100\% data, standard GAN produced an FID score of 1.31. The \xAI{}$_{shap}$, \xAI{}$_{lime}$ and \xAI{}$_{saliency}$ systems resulted in scores of 1.36, 1.21 and 1.26 respectively. The \xAI{}$_{lime}$ system had the best performance and resulted in an improvement of 7.35\% in the FID score, as compared to standard GAN. For 35\% data, standard GAN produced a score of 1.75 while the \xAI{}$_{shap}$, \xAI{}$_{lime}$ and  \xAI{}$_{saliency}$ systems produced scores of 1.50, 1.41 and 1.55 respectively. All three \xAI{} systems outperformed standard GAN in this setting, with the \xAI{}$_{lime}$ system resulting in an improvement of 19.62\%. A sample of the images generated by the \xAI{}$_{lime}$ system using 35\% data can be seen in Figure~\ref{fig:viewMNIST}.

The results of the experiments on the Fashion MNIST dataset can be found in Figure~\ref{fig:results_fmnist}. For 100\% data, standard GAN produced an FID score of 1.16. The \xAI{}$_{shap}$, \xAI{}$_{lime}$ and \xAI{}$_{saliency}$ systems produced scores of 1.06, 0.97 and 1.1 respectively. Again, the \xAI{}$_{lime}$ system had the best results, with an improvement of 16.67\% over standard GAN. For 35\% data, standard GAN produced a score of 1.61. The \xAI{}$_{shap}$, \xAI{}$_{lime}$ and \xAI{}$_{saliency}$ systems produced scores of 1.24, 1.35 and 1.34 respectively. Here, the \xAI{}$_{shap}$ system had the best performance, with an improvement of 23.18\%. A sample of the images generated by the \xAI{}$_{shap}$ system using 35\% data can be seen in Figure~\ref{fig:viewFMNIST}.

The average time taken by each of the GANs on the respective dataset can be found in Table~\ref{tab:times_mnist}. The \xAI{}$_{saliency}$ system runs in about 2x the time that standard GAN does, while the \xAI{}$_{shap}$ and \xAI{}$_{lime}$ systems take around 10x and 35x the time that standard GAN requires respectively. The reason for the discrepancy in times between the \xAI{}s systems is due to the difference between their implementation. Overall, \xAI{} has been shown to outperform standard GAN in terms of FID scores, with improvements of up to 23.18\%.

\begin{figure}[tb]
\centering
\includegraphics[width=0.9\columnwidth]{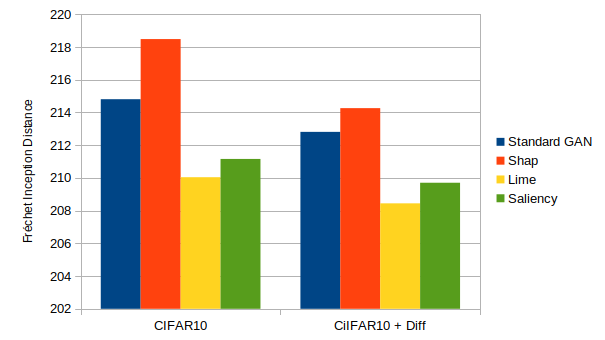}
\caption{FID scores on CIFAR10 Dataset}\label{fig:results_cifar}
\vskip 4mm
\includegraphics[width=0.9\columnwidth]{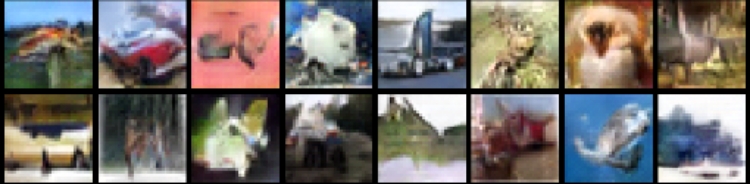}
\caption{Sample of Images Generated by the \xAI{}$_{saliency}$ System using 20\% Data on CIFAR10 Dataset}
\label{fig:view_cifar}
\end{figure}

\subsection{Results on CIFAR10 Dataset}
\label{sec:cifar}

We next ran our experiments on the CIFAR10 dataset using the parameters described earlier. In order to view the efficacy of \xAI{} in the case where data is scarce, we used 100\% of the data for the standard GAN while only using 20\% data for \xAI{}. Further, to compare with the work in~\cite{data_efficient_gan}, we used the Differential Augmentation implementation code linked in the paper to run another set of experiments. In the latter experiments, we added Differential Augmentation to all GANs - which we will hereafter refer to as ``+ Diff". Note that standard GAN+Diff still uses 100\% of the data while all the \xAI{}s+Diff use 20\% data. The results of these experiments are found in Figure~\ref{fig:results_cifar}. In the first run of the experiment (i.e without Differential Augmentation), standard GAN resulted in a FID score of 214.81 while the \xAI{}$_{shap}$, \xAI{}$_{lime}$ and \xAI{}$_{saliency}$ systems resulted in scores of 218.48, 210.04 and 211.16 respectively. The \xAI{}$_{lime}$ system showed the best results with around 2.22\% improvement in FID score over standard GAN, even with 20\% of the data.

Adding Differential Augmentation to standard GAN resulted in an FID score of 212.81, which is 0.93\% improvement in FID score over standard GAN. As previously shown, \xAI{} (in particular the \xAI{}$_{lime}$ system) produced better results than Differential Augmentation - even when using 20\% of the data. Moreover, both methods can be combined to produce complementary results. Running \xAI{}s with Differential Augmentation produced better scores for all GANs, with \xAI{}$_{shap}$+Diff, \xAI{}$_{lime}$+Diff and \xAI{}$_{saliency}$+Diff resulting in scores of 214.27, 208.45 and 209.70 respectively.

The times taken (in seconds) by each of the GANs to run the experiments can be found in Table~\ref{tab:times_cifar}. \xAI{}$_{saliency}$ takes around 2x the time and \xAI{}$_{lime}$ and \xAI{}$_{shap}$ take around 25x the time that standard GAN requires. The time taken by \xAI{}$_{shap}$ system is similar to the \xAI{}$_{lime}$ system as the implementation of shap is expensive in the case of color images. Overall, \xAI{}$_{lime}$+Diff shows an improvement of 2.96\% in FID score over standard GAN while using 20\% of the data. Note that the discrepancy in FID scores (and conversely, the training time) between \xAI{} and standard GAN would be higher if \xAI{}s used 100\% of the dataset for training. 

\subsection{Discussion of Experimental Results}
\label{sec:discussion}
We performed extensive experiments using MNIST, Fashion MNIST and CIFAR10 datasets on both fully connected and DC GANs and showed that \xAI{}, particularly one using the lime xAI system, results in improvements of up to 23.18\% in FID scores over standard GAN. We compared our work with~\cite{data_efficient_gan} and showed that \xAI{} resulted in improvement over Differential Augmentation in our experiments, and that both techniques are complementary and can be combined to result in even more improvement in FID scores. We believe that the important take-away is that \xAI{}s show an improvement over standard GANs in terms of FID score even when using less data.

\noindent\textbf{Regarding the increased training time of \xAI{}:} While \xAI{} requires more training time compared to standard GANs due to the overhead of the xAI system, we believe this is still an advantageous trade-off. GAN research focuses on improving the quality of the images and handling data scarcity, and consequently the time required to train is not as important. In addition, xAI systems scale linearly with the number of neurons in the DNN model. Therefore, the overhead caused by the xAI system will be linear to the model - allowing most hardware that can train standard GANs to be able to train \xAI{}s. Furthermore, with the advent of parallel and distributed computing as well as easier access to powerful computational resources, the time required to train \xAI{} will be further mitigated. 

\section{Future Work}

Our results suggest \xAI{} can be leveraged in settings where data efficiency is important - such as where training data is limited or in privacy conscious settings. It can be also used in normal settings to produce better quality GANs.

\subsection{Controlling How Models Learn}
\label{sec:control}

While standard GANs only use one value (loss) to calculate corrective feedback to the generator, there are many ways this feedback is used. For instance, several GANs vary the type of loss function~\cite{unrolled_gan,wasserstein,energyGAN,mmdgan,wasserstein_improved} and the selection of the optimizer (such as Stochastic Gradient Descent) to control how the model learns. Similarly, we believe that the feedback provided by xAI system using \xAI{} - which is \enquote{richer} compared to only using the loss value - can allow for greater control over this learning process. This control can be applied in various ways, such as in selecting the type of xAI system to use, varying the parameters of the chosen xAI system, offsetting the mask $M$ to adjust the weight given to xAI feedback, alternating between xAI-guided and standard generator training, and selecting methods to combine xAI feedback with loss. We argue that \xAI{}s are a powerful way for users to gain greater control over the training process of GAN models, and that there are many avenues, applications, and extensions of this idea worth exploring in the future. 

\section{Conclusion}

In this paper, we introduce \xAI{}s, a class of generative adversarial network (GAN) that use an explainable AI (xAI) system to provide \enquote{richer} feedback from the discriminator to the generator to enable more guided training and greater control. We next overview xAI systems and standard GAN training and then introduce our xAI-guided generator training algorithm, contrasting it's difference with standard generator training. To the best of our knowledge, \xAI{} is the first GAN to utilize xAI feedback for training. We perform experiments using MNIST and Fashion MNIST datasets and show that \xAI{} has an improvement in Fréchet Inception Distance of up to 23.18\% as compared to standard GANs. In addition, we train \xAI{} on the CIFAR10 dataset using only 20\% of the data and compare it with standard GAN trained on 100\% and show that \xAI{} outperforms standard GAN even in this setting. We compare our work to the Differentiable Augmentation technique and show that \xAI{} trained on 20\% of the data outperforms standard GAN trained with Differential Augmentation. We further combine \xAI{} with Differential Augmentation to produce even better results. There is a trade-off between data-efficiency, training time and quality of images in GANs and our experiments show that \xAI{}$_{saliency}$ provides the best value out of the xAI systems compared. Ultimately, \xAI{} may enable greater control over the GAN learning process - allowing for better performance as well as a better understanding of GAN learning.


\bibliography{sources}
\end{document}